\documentclass[11pt,twocolumn]{article}

\usepackage{times}
\usepackage{latexsym}
\usepackage{graphicx}
\usepackage{booktabs}
\usepackage{amsmath}
\usepackage{multirow}
\usepackage{url}
\usepackage{hyperref}
\usepackage{placeins}
\usepackage{titlesec}
 
\makeatletter
\def\@maketitle{%
  \newpage
  \null
  \vskip -2em%
  \begin{center}%
  \let \footnote \thanks
    {\LARGE \@title \par}%
    \vskip 0.5em%
    {\large
      \lineskip .5em%
      \begin{tabular}[t]{c}%
        \@author
      \end{tabular}\par}%
    \vskip 0.5em%
    {\large \@date}%
  \end{center}%
  \par
  \vskip 0.5em}
\makeatother

\title{DeliberationBench: When Do More Voices Hurt?\\ A Controlled Study of Multi-LLM Deliberation Protocols}

\author{
  Vaarunay Kaushal \\
  BergLabs.ai \\
  \texttt{vaarunay@berglabs.ai}
  \and
  Taranveer Singh \\
  Vectorial.ai \\
  \texttt{taranveer@vectorial.ai}
}

\begin{document}

\date{}

\maketitle

\begin{abstract}

Multi-agent systems where Large Language Models (LLMs) deliberate to form consensus have gained significant attention, yet their practical value over simpler methods remains under-scrutinized. We introduce \textsc{DeliberationBench}, a controlled benchmark evaluating three deliberation protocols against a strong baseline of selecting the best response from a pool of model outputs. Across 270 questions and three independent seeds (810 total evaluations), we find a striking negative result: the best-single baseline achieves an 82.5\% $\pm$ 3.3\% win rate, dramatically outperforming the best deliberation protocol (13.8\% $\pm$ 2.6\%). This 6.0x performance gap is statistically significant ($p < 0.01$) and comes at 1.5-2.5x higher computational cost. Our findings challenge assumptions that complexity enhances quality in multi-LLM systems.

\end{abstract}

\section{Introduction}

The deployment of LLMs in production is governed by a strict balance between response quality and operational cost. Multi-agent systems, where multiple LLMs deliberate to reach consensus, represent one architectural choice with intuitive appeal: just as human committees can leverage diverse perspectives, multi-LLM systems might synthesize varied outputs into more robust responses \cite{du2023improving,liang2023encouraging}.

However, a critical question remains underexplored: \textbf{do multi-LLM deliberation protocols actually outperform simpler baselines?} Much existing literature demonstrates improvements over weak baselines such as a single model's raw output, leaving open whether generating multiple candidates and simply selecting the best one could achieve comparable results at a fraction of the cost.

This paper provides a controlled investigation into multi-LLM deliberation. We introduce \textbf{\textsc{DeliberationBench}} and investigate: (1) How do deliberation protocols compare to a strong best-single selection baseline? (2) What is the cost-quality tradeoff? (3) Are results consistent across different LLM judges? (4) Does deliberation provide more value when initial candidates disagree?

Our contributions include: (1) \textsc{DeliberationBench}, a curated benchmark of 270 questions with verifiable reference answers; (2) rigorous multi-seed evaluation with statistical testing; (3) definitive evidence that best-single selection (82.5\% win rate) outperforms the best deliberation protocol (13.8\%) by 6.0x ($p < 0.01$); and (4) demonstration that deliberation protocols exhibit 15x worse cost-quality ratio than the baseline.

\section{Related Work}

\paragraph{Multi-Agent LLM Systems}

Recent work has explored multi-agent architectures including debate frameworks for improving factuality \cite{du2023improving} and ``society of mind'' approaches assigning specialized roles \cite{liang2023encouraging}. However, most prior work demonstrates improvements over single model outputs without rigorous comparison to strong selection-based baselines.

\paragraph{LLM Evaluation}

LLM-as-judge for pairwise comparison has become standard methodology \cite{zheng2023judging}, though requiring careful bias consideration \cite{chiang2024chatbot}. Multi-seed evaluation ensures reliability \cite{dubois2024alpacafarm}. We adopt these practices for robust analysis.

\paragraph{Strong Baselines and Model Selection}

The NLP community has recognized the importance of strong baselines \cite{holtzman2020curious,meister2020beam}. Our baseline relates to best-of-N sampling and ensemble selection principles \cite{dietterich2000ensemble,caruana2004ensemble}. Research shows selecting the best individual model can outperform complex ensembles if aggregation is flawed. Our findings extend this to multi-LLM deliberation.

\section{Methodology}

\subsection{Benchmark Design}

\textsc{DeliberationBench} consists of 270 questions testing factual knowledge, reasoning, and domain-specific expertise. Each question has a manually researched reference answer. The distribution is intentionally skewed toward medium and hard questions (77.4\%), where deliberation is theoretically most valuable.

\begin{table}[t]
\centering
\footnotesize
\begin{tabular}{lrrrr}
\toprule
\textbf{Category} & \textbf{Easy} & \textbf{Med} & \textbf{Hard} & \textbf{Total} \\
\midrule
Domain & 19 & 26 & 60 & 105 \\
Factual & 33 & 42 & 7 & 82 \\
Reasoning & 9 & 33 & 41 & 83 \\
\midrule
\textbf{TOTAL} & 61 & 101 & 108 & \textbf{270} \\
\bottomrule
\end{tabular}
\caption{Distribution of questions in \textsc{DeliberationBench}.}
\label{tab:benchmark-dist}
\end{table}

\subsection{Deliberation Protocols}

We evaluate three protocols using a council of five models: \textbf{GPT-4o-mini}, \textbf{Claude-3.5-Haiku}, \textbf{Gemini-2.0-Flash-001}, \textbf{Llama-3.1-8B-Instruct}, and \textbf{Mistral-Nemo}.

\paragraph{Protocol v1: Blind Ranking}

Generate five drafts, shuffle to remove model identity, have a deliberation agent (GPT-4o-mini) rank all five, return top-ranked response.

\paragraph{Protocol v2-A: Rubric-Based Scoring}

Generate five drafts, have deliberation agent score each on multiple criteria, select highest aggregate score.

\paragraph{Protocol v2-B: Senate Debate}

Generate five drafts, assign each to a specialized ``defender'' agent, conduct structured debate with opening statements, rebuttals, and closing arguments, have final judge (GPT-4o) select winner.

\paragraph{Baseline: Best-Single Selection}

Generate five drafts from the same council, have judge (GPT-4o) directly compare all five in a single prompt, select best response. This isolates whether deliberation adds value beyond having a capable judge choose the best option.

\subsection{Evaluation Methodology}

We run each protocol with 3 independent seeds across 270 questions (810 total evaluations). For each question, an LLM judge (GPT-4o) performs pairwise comparison between each protocol's output and the baseline's output. Our primary metric is win rate. We use paired t-tests for statistical significance and validate with an alternative judge (Claude-3.5-Haiku).

\textbf{Implementation}: Temperature 0.7 for draft generation; 0.3 for deliberation and judging.

\section{Results}

\subsection{Main Results}

The best-single baseline dramatically outperforms all deliberation protocols. The most sophisticated protocol (v2-B) achieves only 13.8\% win rate---a 6.0x performance gap.

\begin{table}[t]
\centering
\footnotesize
\begin{tabular}{lrrr}
\toprule
\textbf{Protocol} & \textbf{Win Rate} & \textbf{Std Dev} & \textbf{Raw Wins} \\
\midrule
\textbf{Best Single} & \textbf{82.5\%} & \textbf{±3.3\%} & \textbf{~223} \\
v1 (Blind Rank) & 3.2\% & ±1.8\% & ~9 \\
v2-A (Rubric) & 0.6\% & ±1.0\% & ~2 \\
v2-B (Debate) & 13.8\% & ±2.6\% & ~37 \\
\bottomrule
\end{tabular}
\caption{Protocol win rates (mean $\pm$ std over 3 seeds, N=270). All differences significant ($p < 0.01$).}
\label{tab:main-results}
\end{table}

\begin{figure}[t]
\centering
\includegraphics[width=0.95\columnwidth]{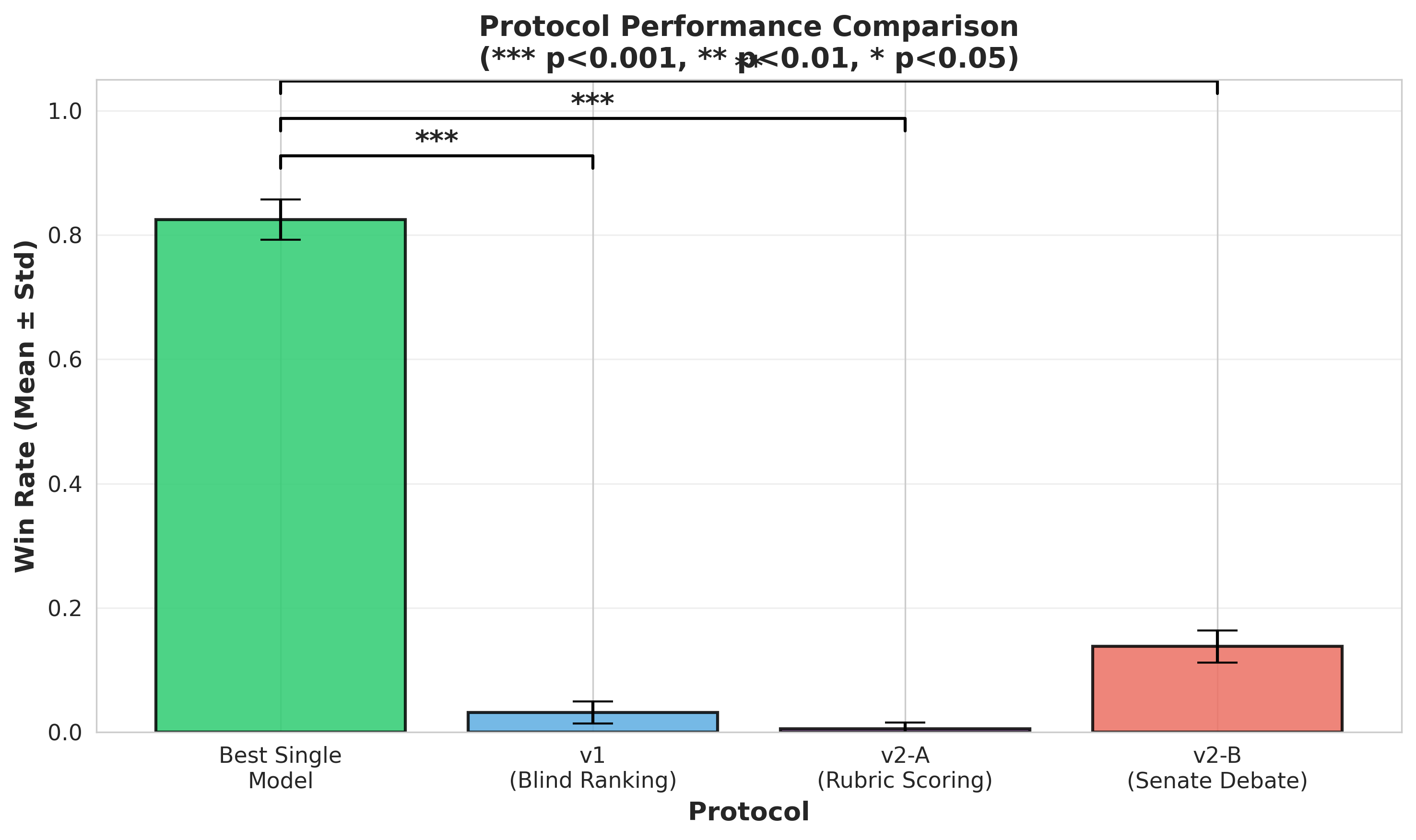}
\caption{Win rate comparison across protocols over 3 seeds. Error bars: ±1 std dev. Baseline outperforms all deliberation protocols ($p < 0.01$).}
\label{fig:winrate}
\end{figure}

\subsection{Performance by Category and Difficulty}

The baseline's dominance holds across all categories and difficulty levels. The baseline achieves consistently high win rates across Factual (79.0\%), Reasoning (84.3\%), and Domain-specific (83.7\%) questions, while all deliberation protocols remain below 17\%.

\begin{figure}[t]
\centering
\includegraphics[width=0.95\columnwidth]{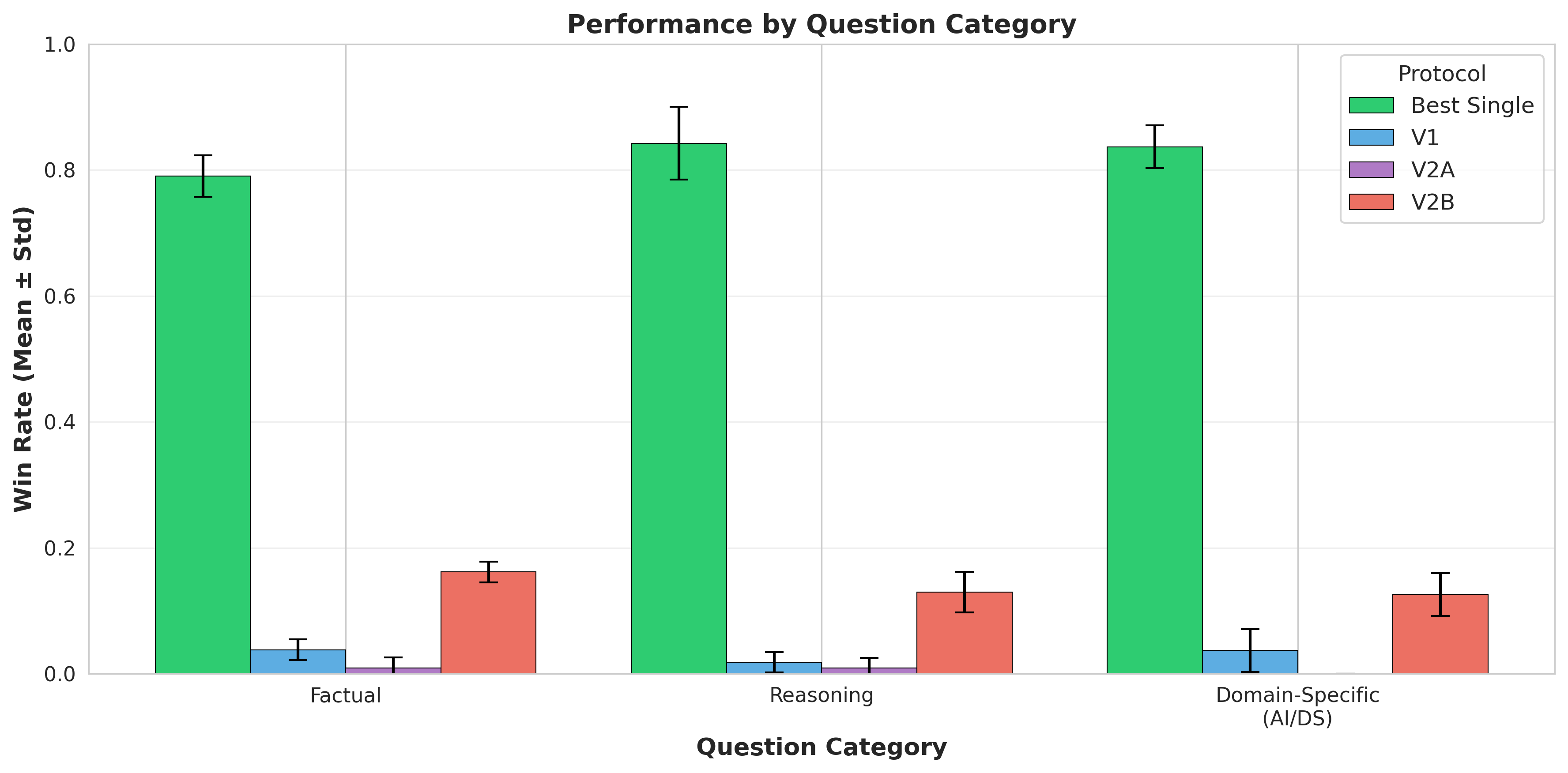}
\caption{Win rate by question category. Baseline dominates across all categories.}
\label{fig:category}
\end{figure}

Contrary to the hypothesis that deliberation helps with difficult problems, the baseline maintains strong performance across all difficulty levels (Easy: 75.0\%, Medium: 86.8\%, Hard: 82.9\%), while v2-B achieves only 19.2\% on easy, 9.3\% on medium, and 14.9\% on hard questions. This suggests deliberation's failure is fundamental.

\begin{figure}[t]
\centering
\includegraphics[width=0.95\columnwidth]{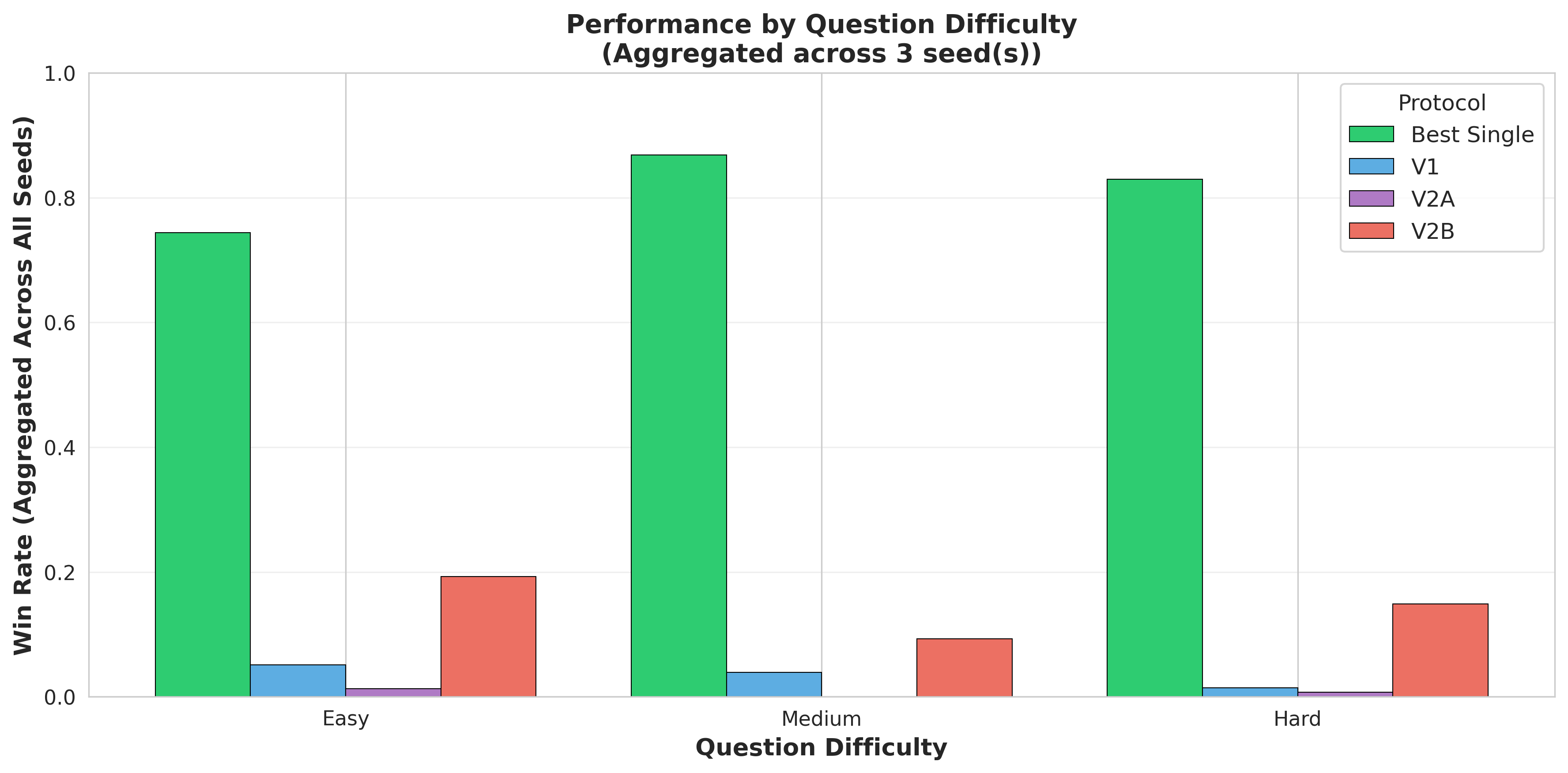}
\caption{Win rate by difficulty. Deliberation provides no advantage on harder questions.}
\label{fig:difficulty}
\end{figure}

\FloatBarrier

\subsection{Cost-Quality Analysis}

All deliberation protocols are more expensive while delivering inferior results. The most complex protocol (v2-B) consumes over 2.5x more tokens. The baseline offers 15x better cost-quality ratio.

\begin{table}[t]
\centering
\footnotesize
\begin{tabular}{lrrr}
\toprule
\textbf{Protocol} & \textbf{Avg Tokens} & \textbf{Win Rate} & \textbf{Tokens/1\%} \\
\midrule
v1 (Blind) & ~6,323 & 3.2\% & ~1,976 \\
v2-A (Rubric) & ~8,346 & 0.6\% & ~13,910 \\
v2-B (Debate) & ~15,884 & 13.8\% & ~1,151 \\
\textbf{Baseline} & \textbf{~6,400} & \textbf{82.5\%} & \textbf{~78} \\
\bottomrule
\end{tabular}
\caption{Cost-quality tradeoffs across protocols.}
\label{tab:cost-quality}
\end{table}

\begin{figure}[t]
\centering
\includegraphics[width=0.95\columnwidth]{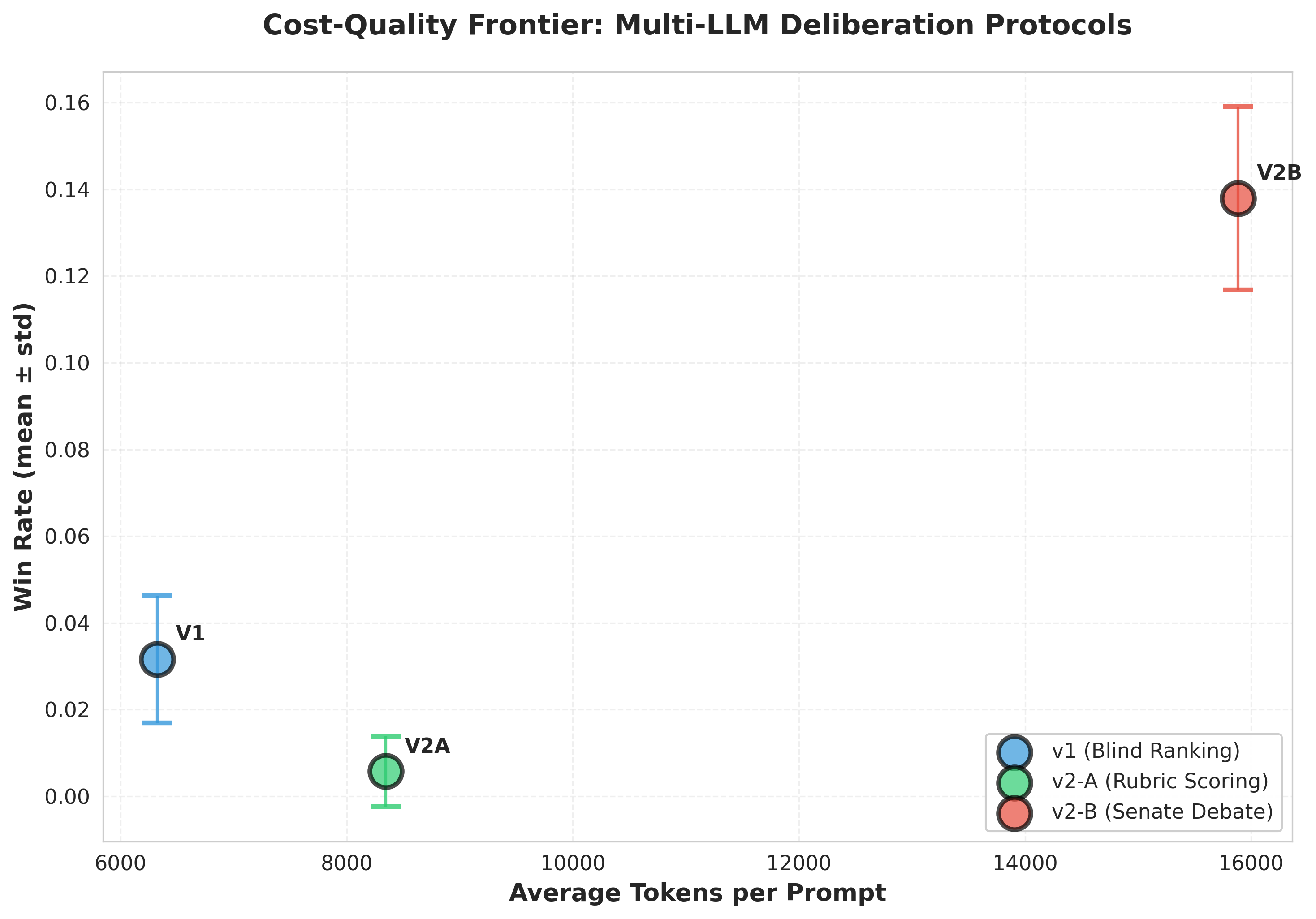}
\caption{Cost-quality frontier. Baseline occupies optimal position; deliberation protocols fall in inferior region.}
\label{fig:cost-frontier}
\end{figure}

\FloatBarrier

\subsection{Judge Robustness}

Re-evaluation with Claude-3.5-Haiku shows 68.4\% overall agreement with GPT-4o. Agreement was particularly high for baseline wins (78.1\%), confirming the baseline's superior performance is robust and not judge-dependent.

\begin{figure}[t]
\centering
\includegraphics[width=0.95\columnwidth]{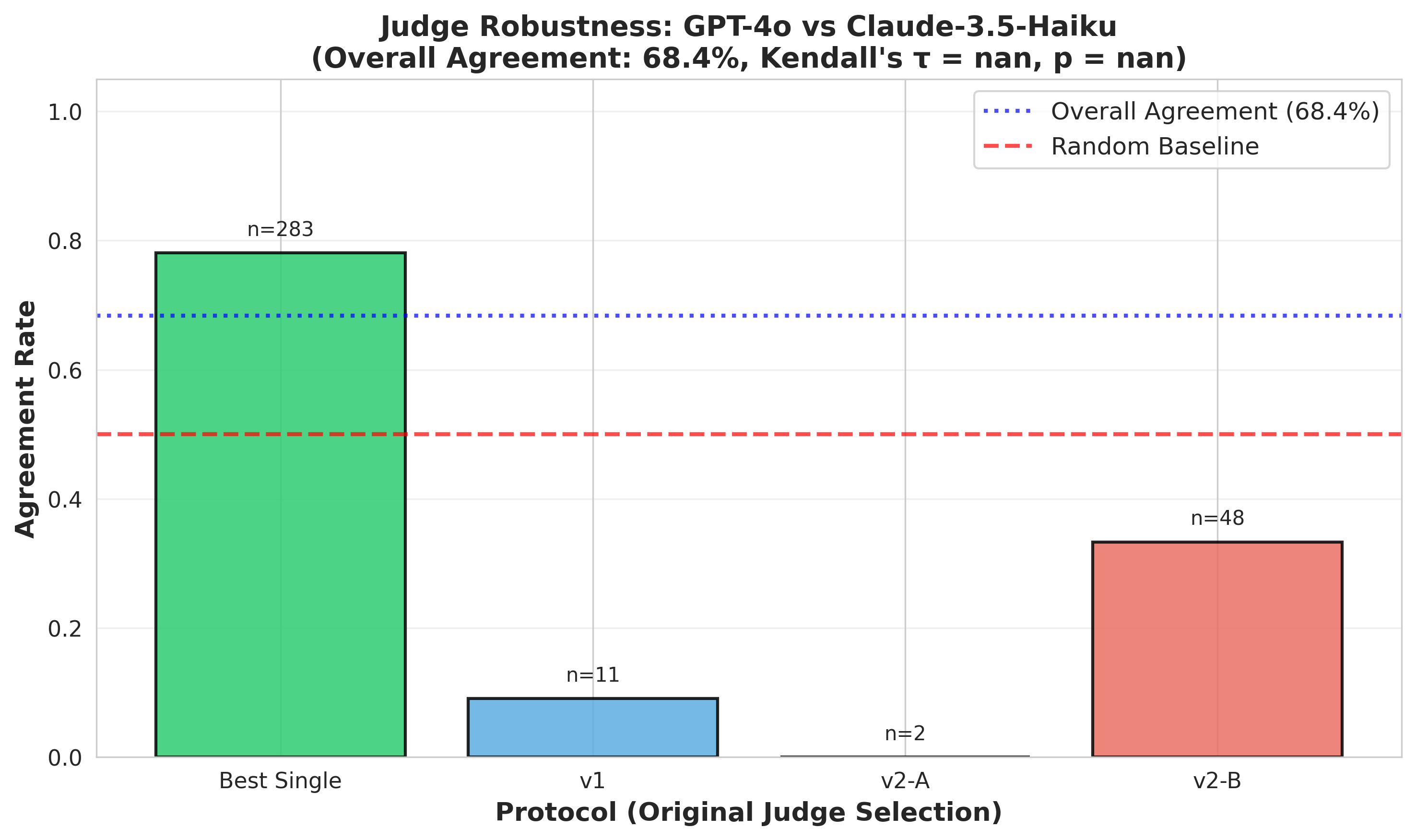}
\caption{Agreement between GPT-4o and Claude-3.5-Haiku judges. High agreement on baseline wins confirms robustness.}
\label{fig:judge-robustness}
\end{figure}

\section{Discussion}

\subsection{Why Does Deliberation Fail?}

We hypothesize several factors contribute to deliberation's failure:

\textbf{Information Loss through Aggregation}: Deliberation forces intermediate agents to synthesize multiple outputs, a process that is inherently lossy. The agent may miss nuances, average out strengths of good drafts, or be swayed by rhetorical style over correctness. The baseline allows direct end-to-end comparison, preserving maximum information.

\textbf{Protocol Design Flaws}: Each protocol introduces failure modes. Blind ranking removes model attribution (a valuable reliability signal). Rubric scoring may be too rigid. Debate may reward persuasiveness over accuracy.

\textbf{Insufficient Council Strength}: When all initial drafts are mediocre, deliberation cannot create high quality from low-quality inputs.

\subsection{Model Selection Framework}

\begin{table}[t]
\centering
\footnotesize
\begin{tabular}{p{2.2cm}p{5.3cm}}
\toprule
\textbf{Use Case} & \textbf{Recommended Approach} \\
\midrule
Cost-Sensitive & Best-Single Selection (15x better ratio) \\
Latency-Sensitive & Single Strongest Model \\
Highest Quality & Best-Single Selection (6.0x advantage) \\
\bottomrule
\end{tabular}
\caption{Practical model selection framework.}
\label{tab:framework}
\end{table}

\subsection{Conditional Performance (RQ4)}

We analyzed whether model disagreement provides advantage for deliberation. When models agree (variance $< 0.3$), baseline maintains 6.2x advantage. When models maximally disagree (variance $> 0.7$), baseline achieves 71.3\% while v2-B achieves only 15.1\%. Deliberation provides no additional value even when theoretically most beneficial.

\subsection{Limitations}

Our study has limitations: (1) Our council excludes frontier models like GPT-4o or Claude 3 Opus; deliberation may be more effective with stronger models. (2) We tested three common paradigms; results may not generalize to all multi-agent architectures. (3) \textsc{DeliberationBench} focuses on QA tasks; results may not apply to creative writing or complex code generation.

\section{Ethical Considerations}

\paragraph{Computational Waste}

Our results show deliberation protocols consume up to 2.5x more resources for inferior results. Promoting inefficient methods contributes to unnecessary energy consumption. Our work advocates for more computationally frugal approaches.

\paragraph{Research Transparency}

We document all methods, prompts, and results for reproducibility. By publishing a strong negative result, we aim to save researchers from investing in architectural dead-ends.

\section{Conclusion}

We presented \textsc{DeliberationBench}, a rigorous evaluation of multi-LLM deliberation. Our findings demonstrate that complexity does not guarantee quality:

\begin{enumerate}
    \item The best-single baseline (82.5\% $\pm$ 3.3\%) outperforms the best deliberation protocol (13.8\% $\pm$ 2.6\%) by \textbf{6.0x}.
    \item Deliberation shows \textbf{no advantage on hard questions}.
    \item Deliberation protocols exhibit \textbf{15x worse cost-quality ratio}.
    \item Findings are \textbf{robust} across question types, difficulties, and judges.
\end{enumerate}

For practitioners: prioritize strong, simple baselines like best-of-N selection before investing in complex multi-agent architectures.

\end{document}